\documentclass[runningheads]{llncs}
\usepackage[T1]{fontenc}
\usepackage{authblk}
\usepackage{tabularx}
\usepackage{amsmath}
\usepackage{amssymb}
\usepackage{graphicx}
\usepackage{verbatim}
\usepackage{booktabs}
\usepackage{changepage}
\usepackage{footnote}

\begin{document}
\title{Deep Graph Neural Point Process For Learning Temporal Interactive Networks}
%\author{First Author\inst{1}\orcidID{0000-1111-2222-3333} \and
%Second Author\inst{2,3}\orcidID{1111-2222-3333-4444} \and
%Third Author\inst{3}\orcidID{2222--3333-4444-5555}}
%\institute{Princeton University, Princeton NJ 08544, USA \and
%Springer Heidelberg, Tiergartenstr. 17, 69121 Heidelberg, Germany
\titlerunning{Deep Graph Neural Point Process For Learning Temporal Interactive Networks}
\author{Su Chen\inst{1,2,3*}\and
Xiaohua Qi\inst{1*} \and Xixun Lin\inst{1**} \and Yanmin Shang\inst{1,2} \and Xiaolin Xu\inst{1,4} \and Yangxi Li\inst{3} }

\makeatletter
\let\@oldmaketitle\@maketitle
\def\@maketitle{%
  \@oldmaketitle%
  \renewcommand{\thefootnote}{}% 
  \footnotetext{*Equal Contribution.}%
  \footnotetext{**Corresponding Author.}%
  
  \renewcommand{\thefootnote}{\arabic{footnote}}% 
}
\makeatother

\authorrunning{S. Chen et al.}
\institute{Institute of Information Engineering, Chinese Academy of Sciences, Beijing, China\\ \email{\{chensu,shangyanmin,linxixun\}@iie.ac.cn, 202100460110@mail.sdu.edu.cn} \and
School of Cyber Security, University of Chinese Academy of Sciences, Beijing, China \and National Computer Network Emergency Response Technical Team/Coordination Center of China, Beijing, China\\ \email{liyangxi@outlook.com}\and Zhongguancun Laboratory, Beijing, China\\ \email{xuxl@zgclab.edu.cn}
}

\maketitle              % typeset the header of the contribution
\begin{abstract}

Learning temporal interaction networks(TIN) is previously regarded as a coarse-grained multi-sequence prediction problem, ignoring the network topology structure influence. This paper addresses this limitation and a \textbf{D}eep \textbf{G}raph \textbf{N}eural \textbf{P}oint \textbf{P}rocess(DGNPP) model for TIN is proposed. DGNPP consists of two key modules: the Node Aggregation Layer and the Self Attentive Layer. The Node Aggregation Layer captures topological structures to generate static representation for users and items, while the Self Attentive Layer dynamically updates embeddings over time. By incorporating both dynamic and static embeddings into the event intensity function and optimizing the model via maximum likelihood estimation, DGNPP predicts events and occurrence time effectively. Experimental evaluations on three public datasets demonstrate that DGNPP achieves superior performance in event prediction and time prediction tasks with high efficiency, significantly outperforming baseline models and effectively mitigating the limitations of prior approaches. 
\end{abstract}

\section{Introduction}
Temporal Interactive Networks(TIN) are those in which the interactions between nodes (representing entities such as people, devices, or systems) change over time. Unlike traditional static networks, TIN, which combine temporal information and interaction modeling, are a useful resource to reflect the relationship between users and items over time, and have been successfully applied to many real-world domains~\cite{r9,r10,r13,r14} such as recommendation systems~\cite{lin2025contrastive,10.1145/3711896.3736951,10.1145/3711896.3737269}, user profiling~\cite{yin2014temporal,lin2021task,lin2023towards}, and social networks~\cite{lin2024graph,cao2025ibpl,lin2025conformal}.

In recent years, despite the relevant research on TIN having been developing continuously, we argue that one important problem remains unsolved, namely the topology structure in TIN is not fully utilized. In previous studies, learning temporal interaction networks is generally regarded as a coarse-grained multi-sequence prediction problem, ignoring the network topology structure influence between the interactive sequences on the prediction results. In fact, sequence features are actually dependent on network structures. For example,In recommendation systems, users' browsing, buying, or viewing behaviors are also influenced by the actions of other users in the social network. By leveraging the network topological dependencies across different sequences, we can predict the next step more precisely.

In this work,we propose a model named \textbf{D}eep \textbf{G}raph  \textbf{Neural} \textbf{P}oint \textbf{P}rocess (DGNPP), incorporating topological structures into dynamically embeddings updating over time with an highly efficient graph structure representation. It outperforms over the previous models.Our principal contributions are threefold:
\begin{enumerate}
    \item We propose a novel DGNPP model based on the temporal point process paradigm. The model combines the topological information based on the Temporal Point Process(TPP) framework.
    \item We introduce two innovative modules in the DGNPP model for efficient information aggregation. The Node Aggregation Layer module captures topological structures to generate static representations for users and items. The Self Attentive Layer module dynamically updates embeddings over time. The two modules make it adequately represented and efficiently calculated for the model effects.
    \item Extensive experiments are conducted on three public standard datasets. The results show that the proposed method achieves consistent and significant improvements over recent competitive baselines.   
\end{enumerate}

\section{Related work}

The main methods of the temporal interactive networks include:

Method based on \textbf{Random Walk}~\cite{r12}. By simulating an object moving gradually in space according to some random rule, the direction and distance of each step are usually random, and the decision of each step depends only on the current state, which is independent of the previous state. This method ignores the statistical characteristic information of the random process.

Method based on \textbf{Recurrent Neural Network}~\cite{r11}. Temporal interactive networks are often combined with temporal models such as RNN and LSTM to capture time dependencies. By combining mechanisms such as LSTM or GRU in the network, the model can maintain and update historical states to capture long- and short-term temporal dependencies.

Method based on \textbf{Temporal Point Process}(TPP)~\cite{r18,nguyen2018continuous,r21,r22,r23,r2,r25,lin2021disentangled,zhang2024neural}. By defining an intensity function, we predict the time and probability of the next instance. The classical TPP models include the Poisson process~\cite{r15}, Hawkes process~\cite{r16}, etc.. Using TPP paradigm to build TIN is a hot research direction, previous models usually build the relationship between nodes and edges in TIN based on graph neural networks(GCN). However, general GCN-based TPP structures is time-costly since its topology is relatively complicated. When training on large-scaling data, it's difficult to satisfy the requirements for general application. It is necessary to design a suitable model to handle the relationships between nodes and edges in graph-structured data with high efficiency in large-scaling data.

\section{Background}
\subsection{Temporal Interaction Networks}

In the time interval (0,T], the temporal interaction network can be represented as G(T)={($u_{i},v_{i},t_{i})|u_{i}\in U,v_{i}\in V,t_{i}\in (0,T]$}, where U denotes the user sets and V denotes the item sets. For ($u_{i},v_{i},t_{i})) \in G(T)$, it indicates that $u_{i}$ interacted with item $v_{i}$ at time $t_{i}$.

\subsection{Temporal Point Process}

The Temporal Point Process (TPP) is a mathematical framework used to model the probability of randomly occurring discrete events over time. Given a time series $T = \{t_1, t_2, \dots, t_n\}, where 0 < t_1 < t_2 < \dots < t_n$, we typically use the intensity function $ \lambda(t)$ to represent a Temporal Point Process. In the time interval [t,t+dt),$\lambda(t)^{*}dt$denotes the probability of an event occurring within 
 t~t+dt,given the history of the first $n$ events, ie.,$P(event in [t,t+dt]|Ht)$,which can also be expressed as:

\begin{align}
\lambda^*(t)\mathrm{d}t \quad
&= \quad\frac{f(t|\mathcal{H}_{t_n})\mathrm{d}t}{1 - F(t|\mathcal{H}_{t_n})}= \quad\frac{\mathbb{P}(t_{n + 1} \in [t, t + \mathrm{d}t]|\mathcal{H}_{t_n})}{\mathbb{P}(t_{n + 1} \notin (t_n, t)|\mathcal{H}_{t_n})}\bigskip\notag\\
&= \quad\frac{\mathbb{P}(t_{n + 1} \in [t, t + \mathrm{d}t], t_{n + 1} \notin (t_n, t)|\mathcal{H}_{t_n})}{\mathbb{P}(t_{n + 1} \notin (t_n, t)|\mathcal{H}_{t_n})}\bigskip\notag\\
&= \quad\mathbb{P}(t_{n + 1} \in [t, t + \mathrm{d}t]|t_{n + 1} \notin (t_n, t), \mathcal{H}_{t_n})\bigskip\notag\\
&= \quad\mathbb{P}(t_{n + 1} \in [t, t + \mathrm{d}t]|\mathcal{H}_{t -})= \quad\mathbb{E}[N([t, t + \mathrm{d}t])|\mathcal{H}_{t -}]\notag
\end{align}

In the above formulation, $H_{t_{n}}$represents the history of the first $n$ events, $N(t,t+dt)$ denotes the number of events occurring in time interval $(t,t+dt)$ (it can be either 0 or 1), and f($t|H_{t_{n}}$) represents the conditional intensity function.

In modeling TPP, the Multivariate Hawkes Process (MHP) plays a crucial role. Given a set of sequences of events $S(T)={(m_{1},t_{1}),(m_{2},t_{2})......(m_{n},t_{n})}$, we assume that the baseline intensity of event $i$ is $\mu_i$, the influence factor of event $i$ on event $j$ is $\alpha_{ij}$ , and the kernel function is $g$. Then, the intensity function can be denoted as:
\begin{align}
\lambda_i(t) = \mu_i + \sum_{j=1}^{d} \sum_{T_j^{(k)} < t} \alpha_{ij} g(t - t_{n})  \notag 
\end{align}

Furthermore, let $\lambda(t)=\sum_{i<M} \lambda_i(t)$ ,which represents the sum of all intensity functions. From the above formulation, it is evident that the function $\lambda_i(t)$  only takes the mutual influence into consideration between events but ignores the topological structure of the temporal interaction network. To address this limitation, our model incorporates Node Aggregation Layer(NAL) architecture, which effectively captures the topological information of the temporal interaction network, thereby enabling more accurate predictions.

\section{Proposed Model}
\subsection{Overall Architecture}

The events occurring within the time interval $(0,T]$ can be abstracted as a temporal interaction network $G(T)={(u_{i},v_{i},t_{i})|u_{i}\in U,v_{i}\in V,t_{i}\in (0,T]}$ where $U$ represents the set of all users and $V$ represents the set of all items that have interacted with users. Consequently, each time a new event occurs, a new edge is added to this temporal interaction network. Thus, predicting the next event can be abstracted as predicting the next edge in the graph.

\begin{figure}[!htb]
\vspace{-10pt}
    \centering
    \includegraphics[width=\linewidth]{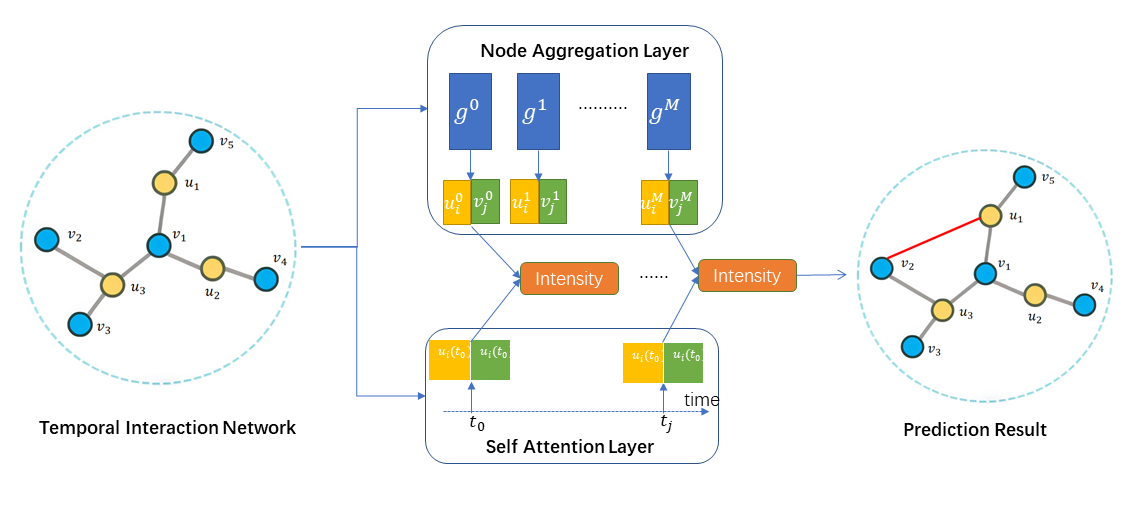}
    \caption{The overall architecture of DGNPP. It consists of two key components:(1) The Node Aggregation Layer,which captures topological structures to generate static representations for users and items. (2) The Self Attentive Layer,which dynamically updates embeddings over time. }
    \vspace{-10pt}
    \label{fig:architecture}
    \vspace{-10pt}
\end{figure}

We select $N$ snapshots to record the temporal interaction network at specific time points $t$, Specifically, we divide the time interval $(0,T]$ into N equal parts, where each segment has a duration of  $d=\dfrac{T}{N}$,We then generate interaction networks at timestamps $0,d, 2d....(N-1)d$. Our model consists of two main modules: the Node Aggregation Layer and the Self Attentive Layer. The Node Aggregation Layer captures the topological information of each temporal interaction network snapshot to generate stable user and item embeddings ($\mathcal{u}$ and $\mathcal{v}$). The Self Attentive Layer is responsible for generating dynamic user and item embeddings (u(t) and v(t)) at any arbitrary timestamp t.
\subsection{Embedding initialization}

We need to initialize the embeddings for the user nodes, item nodes, and time periods. This initialization process consists of two components: the Node Embedding Layer and the Time Embedding Layer.
%\subsubsection{Node Embedding.}

\textbf{Node Embedding.} It serves for converting users and items from one-hot encoding representation into D-dimensional vector representations (where \(u_i \in \mathbb{R}^D\) and \(v_j \in \mathbb{R}^D\)). This conversion is accomplished through the multiplication of the user feature vector\(\mathbf{u}\)and an initialized embedding matrix.
%\subsubsection{Time Embedding.}

\textbf{Time embedding.}Unlike discrete positional encoding, time is an inherently continuous variable. As a result, discrete representations cannot capture the nuances of time interval information. In our proposed model, with the aim of seamlessly integrating continuous temporal information and the sequential information of discrete events, we design the Time Embedding layer. Consider a time series where the x-th event is characterized by the triple \((u_{x}, v_{x}, t_{x})\). For any time instant t such that \(t > t_{x}\), the embedding \(l(t)\) can be computed using the following formula:

\begin{align}
 [{p}_{t^x}(t)]_i=\left\{
\begin{array}{rcl}
\cos(\omega_i(t-t^x)+\frac{h}{10000^{\frac{i-1}{D}}})       &      & {if i is odd}\\
\sin(\omega_i(t-t^x)+\frac{h}{10000^{\frac{i}{D}}})     &      & {if i is even}
\end{array} \right. 
\end{align}

where $[{p}_{t^x}(t)]_{i}$ denotes the value of the embedding at time instant t along the i-th dimension, and \(\omega_i\) is a learnable parameter used to modify the time interval from \(t^x\) to t.

\subsection{Node Aggregation Layer}

In this section, we design the Node Aggregation Layer to learn the representations of node features and better represent the topological structure on the snapshot graph. Given that the attributes of user nodes and item nodes are not rich, we do not adopt the linear transformation operation and the non-linear activation function. Since these two methods not only consume a substantial amount of time but also have no effect on feature learning of nodes.We adopt the method of summing the weights of neighboring nodes. Specifically, let the number of item nodes connected to a certain user node u be \(|\mathcal{N}_u|\), and the number of user nodes connected to a certain item node v be \(|\mathcal{N}_v|\). Then, in the k-th layer, its forward - propagation operation is as follows:

\begin{equation}
    \begin{aligned}
    \mathbf{u}^{k + 1} &= \sum_{v \in \mathcal{N}_u} \frac{1}{\sqrt{|\mathcal{N}_u|}\sqrt{|\mathcal{N}_v|}}\mathbf{v}_i^{k} \\
    \mathbf{v}^{k + 1} &= \sum_{u \in \mathcal{N}_v} \frac{1}{\sqrt{|\mathcal{N}_v|}\sqrt{|\mathcal{N}_u|}}\mathbf{u}_u^{k}.
\end{aligned}
\end{equation}

In the above equation, to prevent the continuous growth of the scale of embeddings during the graph convolution operation,we select $\sqrt{|\mathcal{N}_v|}\sqrt{|\mathcal{N}_u|}$ as the normalization coefficient. After obtaining the embedding representations from the 0-th layer to the R-th layer for users, denoted as \(u^0, u^1, \ldots, u^R\), and those for items, denoted as\(v^0, v^1, \ldots, v^R\), we perform a combination operation on these embeddings to obtain the final stable user/item embeddings:

\begin{align}
   \mathbf{u} = \frac{1}{R+1}\sum_{r = 0}^{R} \mathbf{u}^{(r)} \quad \mathbf{v} =\frac{1}{R+1} \sum_{r = 0}^{R} \mathbf{v}^{(r)} 
\end{align}

Although the weights of each layer can be set as learnable parameters, to simplify the model, we set the weights of the embeddings for each layer to be the same value of\(\frac{1}{R + 1}\)

\subsection{Self Attentive Layer}

In this section, we introduce the self-attention mechanism to capture the long-range dependencies among different snapshots. Previous studies typically employed models such as RNNs and LSTMs to obtain this. However, these methods are not good at handling long sequences, which may lead to the loss of dependencies on the graph topological structure. Nevertheless, due to the incorporation of the self-attention mechanism, our Self Attentive Layer(SAL) module can more accurately reflect the historical changes in the interaction information of the temporal graph network at different moments.

The Self Attentive Layer is further divided into two components, namely the Attentive Interaction Layer and the Temporal Shift Layer. The Attentive Interaction Layer incorporates temporal information and node information, enabling to generate a new time-related embedding representation of the interaction information\((u_i, v_i, t_i)\). The Temporal Shift Layer outputs the dynamic embeddings of users and items at any given moment based on the embedding information obtained from the previous layer.

\textbf{Attentive Interaction Layer.} It takes an embedding for each event\((u_i, v_i, t_i)\). We need to take into account user information, item information, and temporal information. The temporal information includes both the moment when the event occurs and the order in which the event takes place. The calculation formula is as follows:

\begin{align}
    \Theta_m = \left(Q[\boldsymbol{u}_i|(\boldsymbol{v}_i + \boldsymbol{p}_{t^{|H_{u_{i}}|}}(t))]\right)^{\top} K[\boldsymbol{u}_i|(\boldsymbol{v}^m + \boldsymbol{p}_{t^m}(t))],\text{where } (v^m, t^m) \in \mathcal{G}_{u_m}(t) \notag
\end{align}
\begin{align}
    \alpha_m = \frac{\exp(\Theta_m)}{\sum_{(v^n,t^n)\in\mathcal{G}_{u_i}(t)} \exp(\Theta_n)}, 
\end{align}
\begin{align}
    \boldsymbol{w}^t = ReLU\left(\sum_{(v^n,t^n)\in \mathcal{G}_{u_i}(t)} \alpha_n V \boldsymbol{v}^n\right)\notag
\end{align}

where \(\mathcal{G}_{u_m}(t)\)represents the interaction sequence of user \(u_m\) prior to timestamp t, and \(|\mathcal{G}_{u_m}(t)|\) is the number of elements in the interaction sequence. Q, K, V are the query, key, and value matrices under the self-attention mechanism. Finally, we calculate the embeddings of users and items at time t based on the GRUs model:
\begin{equation}
    \begin{aligned}
    \boldsymbol{u}_i(t)=g_u(\boldsymbol{u}_i,\boldsymbol{w}^t),\\
    \boldsymbol{v}_j(t)=g_v(\boldsymbol{v}_j,\boldsymbol{w}^t)
    \end{aligned}
\end{equation}

%\subsubsection{Temporal Shift Layer}

\textbf{Temporal Shift Layer.}In the layer described above, embeddings were generated only for users and items at a specific moment in time. We now extend this approach to cover the entire time period $(0,T)$. 

To capture this evolution, we introduce a learnable temporal shift layer, which uses the time difference ($t^+-t$) to model the passage of time. The calculation for dynamic embeddings is as follows:

\begin{equation}
    \begin{aligned}
    \boldsymbol{u}_i(t^{+})=(1 + (t^+-t)\boldsymbol{w}_{u_i}) * \boldsymbol{u}_i(t),\\
    \boldsymbol{v}_j(t^{+})=(1 + (t^+-t)\boldsymbol{w}_{v_j}) * \boldsymbol{v}_j(t)
    \end{aligned}
\end{equation}

Through the above formula, at any given time  $t^+$, there is a user/item embedding that evolves continuously over time.
\subsection{Model Training}

In proposed model, stable user and item embeddings are generated through the Node Aggregation Layer, while dynamic user and item embeddings are produced by the Self Attentive Layer. We design a method that incorporates both dynamic and static embeddings into the event intensity function $\lambda(t)$. Next, we optimize our model using maximum likelihood estimation, and the optimized model is used to predict both event types and occurrence time across two dimensions.

\textbf{Intensity Function.} In our model, for any given user-item pairs, a conditional intensity function $\lambda_{(u_i,v_i)}(t)$ is generated. Using the previously generated static and dynamic embeddings, we compute a numerical value for the user-item interaction. This value is then passed through a softmax function to ensure each intensity function produces a positive real number.
\begin{equation}
    \begin{aligned}
        \lambda_{(u_i, v_j)}^{'}(t)=\underbrace{(\boldsymbol{u}_i^{\lfloor\frac{t}{d}\rfloor})^{\top} \boldsymbol{v}_j^{\lfloor\frac{t}{d}\rfloor}}_{\text{static intensity}}+\underbrace{(\boldsymbol{u}_i(t))^{\top} \boldsymbol{v}_j(t)}_{\text{dynamic change}}
    \end{aligned}
\end{equation}

\begin{align}
    \lambda_{(u_i, v_j)}(t)=\frac{exp(\lambda_{(u_i, v_j)}^{'}(t))}{\sum_{(u_i,v_i)\in( 
    \boldsymbol{U},\boldsymbol{V})}exp(\lambda_{(u_i, v_j)}^{'}(t))}
\end{align}

In the expression on the left side, $\lfloor\frac{t}{d}\rfloor$is the index of the graph snapshot generated in the Node Aggregation Layer. This expression remains constant within a time interval of $\frac{T}{N}$, therefore it can be analogized to the baseline intensity $\mu$ in Hawkes process. On the right side the expression is derived from the dynamic embeddings generated by SAL, which change continuously over time. This can be likened to the influence function in Hawkes process.

\textbf{Objective Function.} We optimize the model parameters using maximum likelihood estimation, aiming to maximize the probability of the temporal interaction sequence $(u_1,v_1,t_1),(u_2,v_2,t_2),......(u_n,v_n,t_n)$ occurring during the time interval $(0,T]$. The log-likelihood of the sequence can be expressed as follows:

\begin{align}
    \ell =\underbrace{ \sum_{(u_i,v_i ,t_i)\in G(T)} \log \lambda_{(u_i,v_i)}(t_i)}_{ \text{A}} - \underbrace{\int_{t = 0}^{T} \lambda(t)dt}_{\text{ B}}
\end{align}

where $\lambda(t)$ represents the sum of all $\lambda_{(u_{i,v_i)}(t_i)}$, which is the intensity function for the probability of any event occurring over a given time period. We can intuitively understand this objective function as follows:

A represents the logarithm of the product of the intensity functions for all events. Therefore, the larger the value of A, the higher the probability of the event sequence $(u_1,v_1,t_1),(u_2,v_2,t_2),......(u_n,v_n,t_n)$occurring within the time interval $(0,T]$. B represents the logarithmic probability of no events occurring during the time interval $(0,T]$.

\begin{comment}

\end{comment}

\section{Experiment}
\subsection{Prediction Tasks}
%我们的模型可以对下一个事件的事件类型以及事件发生的时间进行预测
Our model can predict both the next event and the time at which it will occur.

%事件预测：问题可以描述为，在给定事件历史序列$H(t)$和用户$u_i$，在$t^+(t^+>t)$时刻下哪个事件类型最有可能发生？这个问题我们通过计算该时刻下的各个事件类型发生的强度函数来计算，即选择强度函数最大的哪个事件：

\textbf{Event Prediction.} The problem can be described as follows: Given the historical event sequence $H(t)$ and a user $u_i$, which event is most likely to occur at time $t^+(t^+>t)$? We approach this problem by calculating the intensity function for each event at the specific time and selecting the event with the largest intensity function:
\begin{align}\label{eventpredict}
    \operatorname{argmax}_{v_j} \frac{\lambda_{(u_i,v_j)}(t^+)}{\sum_{v_c \in \mathcal{V}} \lambda_{(u_i,v_c)}(t^+)}
\end{align}

\textbf{Time Prediction.} The problem can be described as follows: Given the historical event sequence $H(t)$
, a user$u_i$ , and an item $v_j$,at what time t(where $t^+(t^+>t)$) is it most likely that user will interact with item for the last time? We approach this problem by calculating the expected value of the probability density function $f_{(u_i,v_j)}$ over the time period $(t^+,\infty)$:

\begin{align} \label{timepredict}
    time = \int_{t^+}^{\infty} (t^+ - t)f_{(u_i,v_j)}(t)\mathrm{d}t^+    
\end{align}
\subsection{Datasets}

To compare the performance of different models, we tested our model on three benchmark datasets: Reddit, Wikipedia, and MOOC\footnote{http://snap.stanford.edu/jodie}. These datasets are available on the JODIE platform. \begin{comment}The detailed information for these datasets is provided in Table \ref{tab1}.\end{comment} 
There are significant differences in the number of users, items, and interactions among these datasets, which allows us to effectively evaluate whether our model can accurately predict events and time in different scenarios.
\begin{comment}
    \begin{table}[h]
    \centering
    \caption{Datasets Statistics}
    \setlength{\tabcolsep}{7mm}{
    \begin{tabular}{cccc}
    \hline
    \textbf{Datasets} & \(\vert \mathcal{U} \vert\) & \(\vert \mathcal{V} \vert\) & \textbf{Interactions} \\
    \hline
    Reddit     & 10,000 & 985 & 672,447  \\
    Wikipedia  & 8,227  & 1,001 & 157,474  \\
    Last.FM    & 980  &  1,001 & 1,293,103 \\
    \hline
    \end{tabular}
    }
    \label{tab1}
\end{table}
\end{comment}

\subsection{Metrics}

To evaluate the performance of our model, we predict events and time based on formula \ref{eventpredict} and \ref{timepredict}, respectively. For the event prediction task, we generate the intensity function for each event and obtain the ranking information of the correct item. We then evaluate the model performance by using the MRR and Recall@20 metrics: the higher the values of MRR and Recall@20, the better the model's performance. For the time prediction task, we use the RMSE metric for evaluation: the smaller the RMSE value, the better the model's performance.
\subsection{Baselines}

We select the following models as baselines for comparison.\textbf{CTDNE}~\cite{nguyen2018continuous} is a general framework for incorporating temporal information into network embedding methods.\textbf{JODIE}~\cite{r21} is a coupled recurrent neural network that learns the embedding trajectories of users and items not only generating when users take actions but also modeling the future trajectory.\textbf{DGCF}~\cite{r22} is a framework leveraging dynamic graphs to capture collaborative and sequential relations of both items and users. \textbf{DeepCoevolve}~\cite{r23} is a model to learn user and item features based on their mutual interaction RNN graph by defining the intensity function in point processes. \textbf{DSPP}~\cite{r2} is a model for learning temporal interaction network which incorporates a highly complicated topological structure in the model. \textbf{MetaTPP}~\cite{r25} is a meta learning framework where each sequence is treated as a different task.
    
\subsection{Parameters Setting}

DGNPP selects the same embedding size of 128 as the baseline models. The learning rate is chosen from \{0.01,0.001,0.0001,0.00001\}, and the weight decay is selected from \{0.0001,0.00001,0.000001\}. The dropout probability is set to 0.7, and the number of training epochs is set to 100. Both the Node Aggregation Layer and the Self Attentive Layer are set to 4 layers.
\subsection{Experimental Results}
\subsubsection{Event Prediction.}

We assess the model performance using Recall@10 and MRR, with the experimental results presented in Table 1. The results indicate that our model achieves outstanding overall performance. Specifically, on the Last.FM dataset, it ranks first in all metrics, outperforming the competitive baselines by 16.0\% in Recall@10 and 8.1\% in MRR. This highlights our model’s superior ability to capture topological information within the graph, leading to enhanced performance on complicated interaction sequences. On the Wikipedia dataset, our model falls behind of the best baseline by 0.7\% in Recall@10 but maintaining advantage in MRR. On the Reddit dataset, our model surpasses the best-performing baseline by 1.49\% in Recall@10 and achieves a tie with DGCF for the highest MRR score.

\begin{table}[h]
\vspace{-10pt}
\label{tab1}
\centering
\caption{Performance (\%) comparison of item prediction task}
\setlength{\tabcolsep}{2mm}{
\begin{tabular}{c c c c c c c}
\hline
\textbf{Model} & \multicolumn{2}{c}{\textbf{Last.FM}} & \multicolumn{2}{c}{\textbf{Wikipedia}} & \multicolumn{2}{c}{\textbf{Reddit}} \\ 
 & Recall@10 & MRR & Recall@10 & MRR & Recall@10 & MRR \\ \hline
CTDNE & 1.0 & 1.0 & 5.6 & 3.5 & 25.7 & 16.5 \\ 
JODIE & 38.7 & 23.9 & 82.1 & 74.6 & 85.1 & 72.4 \\
DGCF & 45.6 & 32.1 & 85.2 & 78.6 & 85.6 & \textbf{72.6} \\ 
DSPP & 41.2 & 23.4 & \textbf{87.8} & 76.8 & 86.0 & 71.9\\
MetaTPP & 32.5 & 22.7 & 78.6 & 72.0 & 84.8 & 69.8 \\ \hline
DGNPP & \textbf{52.9} & \textbf{34.7} & 87.2 & \textbf{78.9} & \textbf{87.3} & \textbf{72.6} \\ \hline
\end{tabular}
}
\vspace{-10pt}
\end{table}

\textbf{Time Prediction.} Table \ref{tab2} presents a comparison of the performance and efficiency between our model and the competitive models for time prediction tasks. As shown in the chart, our model performs significantly better in time prediction tasks. Notably, on the Last.FM dataset, our model outperforms DSPP by 30.4\% and DeepCoevolve by 37.1\%, respectively. On the Wikipedia dataset, our model outperforms DSPP by 28.8\% and DeepCoevolve by 36.4\%, respectively. On the Reddit dataset, our model outperforms DSPP by 13.3\% and DeepCoevolve by 23.5\%, respectively. Due to the linear operations performed on node features in the Node Aggregation Layer, our model achieves a notable improvement in time prediction tasks with high efficiency compared to the DSPP and DeepCoevolve model, where complicated feature transformation and nonlinear activation are carried out during the node aggregation process.

We ran our model 10 times on each dataset and calculated the average time required to complete one epoch. As shown in Figure 2, our model's runtime on the Last.FM, Wikipedia, and Reddit datasets was reduced by 12.9\%, 16.2\%, and 25.1\%, respectively, compared to the most competitive DSPP model.
\begin{table}[htbp]
\vspace{-5pt}
    \centering
    \caption{Performance (RMSE) comparison of time prediction}
    \label{tab2}
    \setlength{\tabcolsep}{7mm}{
    \begin{tabular}{cccc}
    \hline
    Model & Last.FM & Wikipedia & Reddit \\
    \hline
    DeepCoevolve& 0.62 & 0.66 & 0.68\\
    DSPP & 0.56 & 0.59 & 0.60 \\
    DGNPP & \textbf{0.39} & \textbf{0.42} &\textbf{0.52}\\
    \hline
    \end{tabular}
    }
    \vspace{-10pt}
\end{table}

\subsubsection{Ablation Experiment}

In this section, we conduct experiments on different model variants of DGNPP on the Reddit, Wikipedia, and Last.FM datasets. We removed the Node Aggregation Layer and the Self Attentive Layer from the model respectively, and then readjusted the parameters. The results are shown in Figure 3. We found that the removal of both the Node Aggregation Layer and the Self Attentive Layer has a certain degree of negative impact on the performance of the model, where the decline is most significant on the Reddit dataset. After removing the Node Aggregation Layer, the MRR decreased by 2.9\%, indicating that the proposed layer can effectively capture the topological sequence information in the graph. While after removing the Self Attentive Layer, the MRR decreased by 5.7\%, demonstrating that the Self Attentive Layer plays a crucial role in capturing long-term sequence information.

\begin{figure}[htbp]
\begin{minipage}[t]{0.45\linewidth}
\centering
\includegraphics[height=5cm,width=5cm]{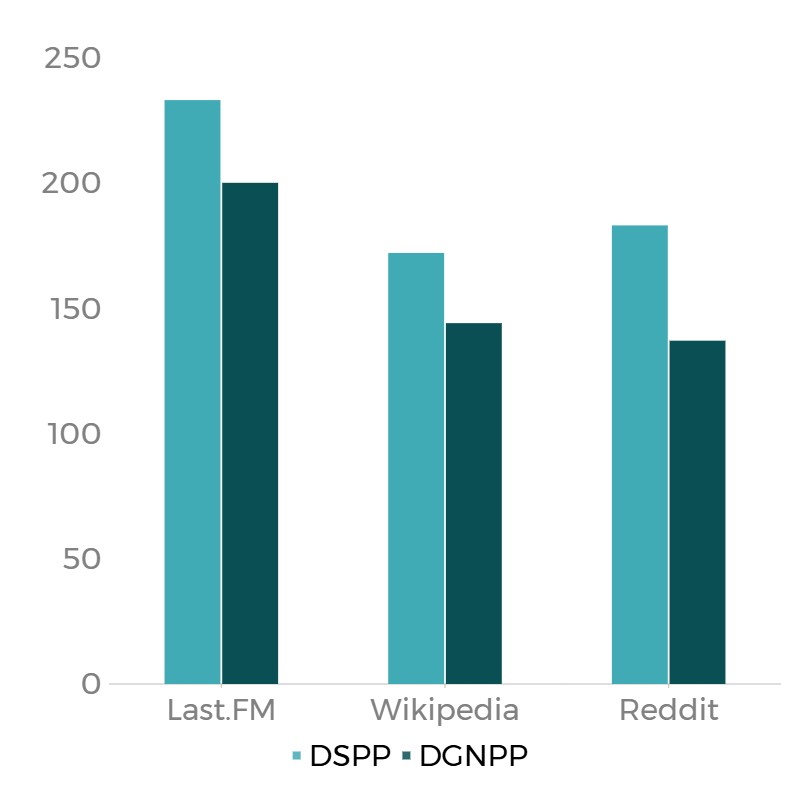}
\caption{Running time comparison}
\end{minipage}%
\begin{minipage}[t]{0.45\linewidth}
\centering
\includegraphics[height=5cm,width=5cm]{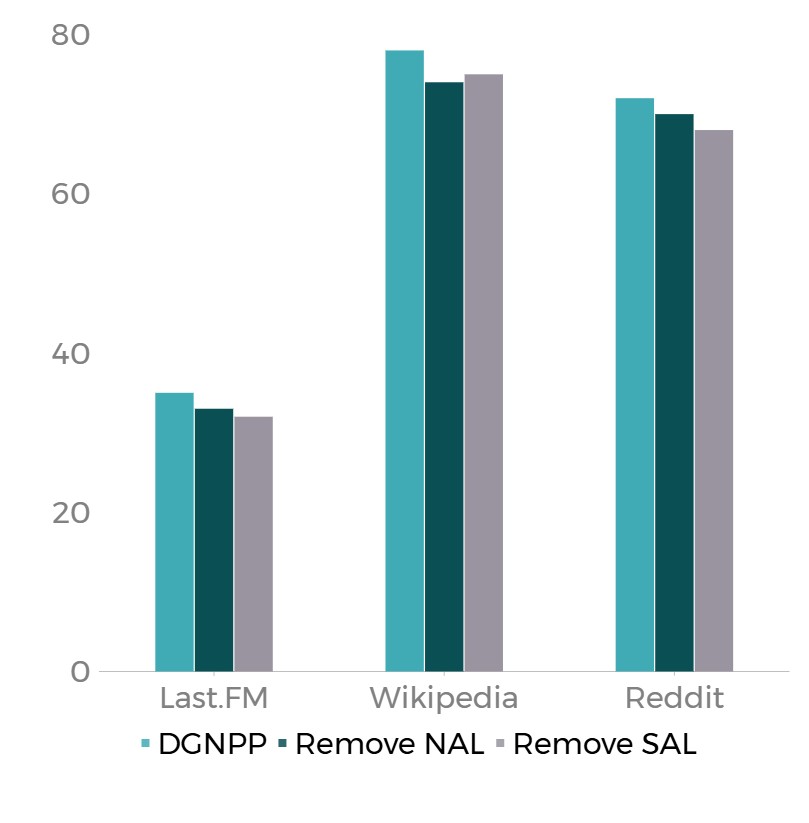}
\caption{Comparison of MRR values of several model variants}
\end{minipage}
\vspace{-10pt}
\end{figure}

\section{Conclusion}

In this paper, to address the limitations of previous models which lacks topological structures or which designs complicated topology designs with inefficiency, we propose a highly efficient model called DGNPP. It consists of two modules: the Node Aggregation Layer(NAL) and the Self-Attentive Layer(SAL). The NAL module is designed to learn the feature representations of nodes in order to capture the topological structure on temporal graph snapshots, while the SAL module aims to learn long-range dependencies within the sequence. Compared with the previous competitive models, our model use an easier and more efficient graph representation with a high training speed. Ultimately, our model demonstrates excellent performance on the Reddit, Wikipedia, and MOOC datasets.

\section*{Acknowledgments}
This work is supported by the Natural Science Foundation of China (No.62192785 and No.62402491) and the National Key Research and Development Program of China (No.2022YFB3102200).

\bibliographystyle{unsrt}
\bibliography{Reference}

\begin{thebibliography}{10}
\providecommand{\url}[1]{#1}
\csname url@samestyle\endcsname
\providecommand{\newblock}{\relax}
\providecommand{\bibinfo}[2]{#2}
\providecommand{\BIBentrySTDinterwordspacing}{\spaceskip=0pt\relax}
\providecommand{\BIBentryALTinterwordstretchfactor}{4}
\providecommand{\BIBentryALTinterwordspacing}{\spaceskip=\fontdimen2\font plus
\BIBentryALTinterwordstretchfactor\fontdimen3\font minus \fontdimen4\font\relax}
\providecommand{\BIBforeignlanguage}[2]{{%
\expandafter\ifx\csname l@#1\endcsname\relax
\typeout{** WARNING: IEEEtran.bst: No hyphenation pattern has been}%
\typeout{** loaded for the language `#1'. Using the pattern for}%
\typeout{** the default language instead.}%
\else
\language=\csname l@#1\endcsname
\fi
#2}}
\providecommand{\BIBdecl}{\relax}
\BIBdecl

\bibitem{r1}
O.~Shchur, A.~C. T{\"u}rkmen, T.~Januschowski, and S.~G{\"u}nnemann, ``Neural temporal point processes: A review,'' \emph{arXiv preprint arXiv:2104.03528}, 2021.

\bibitem{r2}
J.~Cao, X.~Lin, X.~Cong, S.~Guo, H.~Tang, T.~Liu, and B.~Wang, ``Deep structural point process for learning temporal interaction networks,'' in \emph{Machine Learning and Knowledge Discovery in Databases. Research Track: European Conference, ECML PKDD 2021, Bilbao, Spain, September 13--17, 2021, Proceedings, Part I 21}.\hskip 1em plus 0.5em minus 0.4em\relax Springer, 2021, pp. 305--320.

\bibitem{r3}
T.~Gracious, A.~Gupta, and A.~Dukkipati, ``Neural temporal point process for forecasting higher order and directional interactions,'' \emph{arXiv preprint arXiv:2301.12210}, 2023.

\bibitem{r4}
T.~Gracious and A.~Dukkipati, ``Dynamic representation learning with temporal point processes for higher-order interaction forecasting,'' in \emph{Proceedings of the AAAI conference on artificial intelligence}, vol.~37, no.~6, 2023, pp. 7748--7756.

\bibitem{r5}
X.~He, K.~Deng, X.~Wang, Y.~Li, Y.~Zhang, and M.~Wang, ``Lightgcn: Simplifying and powering graph convolution network for recommendation,'' in \emph{Proceedings of the 43rd International ACM SIGIR conference on research and development in Information Retrieval}, 2020, pp. 639--648.

\bibitem{r6}
C.~Wang, O.~Tsepa, J.~Ma, and B.~Wang, ``Graph-mamba: Towards long-range graph sequence modeling with selective state spaces,'' \emph{arXiv preprint arXiv:2402.00789}, 2024.

\bibitem{r7}
T.~Gracious and A.~Dukkipati, ``Dynamic representation learning with temporal point processes for higher-order interaction forecasting,'' in \emph{Proceedings of the AAAI conference on artificial intelligence}, vol.~37, no.~6, 2023, pp. 7748--7756.

\bibitem{r8}
G.~Cencetti, F.~Battiston, B.~Lepri, and M.~Karsai, ``Temporal properties of higher-order interactions in social networks,'' \emph{Scientific reports}, vol.~11, no.~1, p. 7028, 2021.

\bibitem{r9}
D.~Zhou, L.~Zheng, J.~Han, and J.~He, ``A data-driven graph generative model for temporal interaction networks,'' pp. 401--411, 2020.

\bibitem{r10}
S.~Kumar, X.~Zhang, and J.~Leskovec, ``Predicting dynamic embedding trajectory in temporal interaction networks,'' in \emph{Proceedings of the 25th ACM SIGKDD international conference on knowledge discovery \& data mining}, 2019, pp. 1269--1278.

\bibitem{r11}
T.~Omi, K.~Aihara \emph{et~al.}, ``Fully neural network based model for general temporal point processes,'' \emph{Advances in neural information processing systems}, vol.~32, 2019.

\bibitem{r12}
G.~F. Lawler and V.~Limic, \emph{Random walk: a modern introduction}.\hskip 1em plus 0.5em minus 0.4em\relax Cambridge University Press, 2010, vol. 123.

\bibitem{r13}
Q.~Chen, F.~Jiang, X.~Guo, J.~Chen, K.~Sha, and Y.~Wang, ``Combine temporal information in session-based recommendation with graph neural networks,'' \emph{Expert Systems with Applications}, vol. 238, p. 121969, 2024.

\bibitem{r14}
S.~Zhang, L.~Chen, C.~Wang, S.~Li, and H.~Xiong, ``Temporal graph contrastive learning for sequential recommendation,'' in \emph{Proceedings of the AAAI Conference on Artificial Intelligence}, vol.~38, no.~8, 2024, pp. 9359--9367.

\bibitem{r15}
J.~F.~C. Kingman, \emph{Poisson processes}.\hskip 1em plus 0.5em minus 0.4em\relax Clarendon Press, 1992, vol.~3.

\bibitem{r16}
P.~J. Laub, T.~Taimre, and P.~K. Pollett, ``Hawkes processes,'' \emph{arXiv preprint arXiv:1507.02822}, 2015.

\bibitem{r17}
J.~M{\o}ller, A.~R. Syversveen, and R.~P. Waagepetersen, ``Log gaussian cox processes,'' \emph{Scandinavian journal of statistics}, vol.~25, no.~3, pp. 451--482, 1998.

\bibitem{r18}
L.~Zhao, Y.~Song, C.~Zhang, Y.~Liu, P.~Wang, T.~Lin, M.~Deng, and H.~Li, ``T-gcn: A temporal graph convolutional network for traffic prediction,'' \emph{IEEE transactions on intelligent transportation systems}, vol.~21, no.~9, pp. 3848--3858, 2019.

\bibitem{r19}
A.~Hawkes, ``Spectra of some self-exciting and mutually exciting point processes,'' 1971.

\bibitem{r21}
L.~J. Kumar~S., Zhang~X., ``Predicting dynamic embedding trajectory in temporal interaction networks.'' \emph{KDD}, 2019.

\bibitem{r22}
X.~Li, M.~Zhang, S.~Wu, Z.~Liu, L.~Wang, and P.~S. Yu, ``Dynamic graph collaborative filtering.'' \emph{ICDM}, 2020.

\bibitem{r23}
H.~Dai, Y.~Wang, R.~Trivedi, and L.~Song, ``Deep coevolutionary network: Embedding user and item features for recommendation.'' \emph{ArXiv}, 2016.

\bibitem{r24}
J.~Cao, X.~Lin, X.~Cong, S.~Guo, H.~Tang, T.~Liu, and B.~Wang, ``Deep structural point process for learning temporal interaction networks,'' in \emph{European Conference on Machine Learning and Principles and Practice of Knowledge Discovery in Databases (ECML/PKDD)}, 2021.

\bibitem{r25}
W.~Bae, M.~O. Ahmed, F.~Tung, and G.~L. Oliveira, ``Meta temporal point processes,'' in \emph{The International Conference on Learning Representations (ICLR)}, 2023.

\bibitem{r26}
A.~Reinhart, ``A review of self-exciting spatio-temporal point processes and their applications,'' \emph{Statistical Science}, 2018.

\bibitem{nguyen2018continuous}
G.~H. Nguyen, J.~B. Lee, R.~A. Rossi, N.~K. Ahmed, E.~Koh, and S.~Kim, ``Continuous-time dynamic network embeddings,'' in \emph{Companion proceedings of the the web conference 2018}, 2018, pp. 969--976.

\bibitem{zhang2024neural}
S.~Zhang, C.~Zhou, Y.~Liu, P.~Zhang, X.~Lin, and Z.-M. Ma, ``Neural jump-diffusion temporal point processes,'' in \emph{International Conference on Machine Learning}, 2024.

\bibitem{lin2025contrastive}
X.~Lin, R.~Liu, Y.~Cao, L.~Zou, Q.~Li, Y.~Wu, Y.~Liu, D.~Yin, and G.~Xu, ``Contrastive modality-disentangled learning for multimodal recommendation,'' \emph{ACM Transactions on Information Systems}, 2025.

\bibitem{lin2021disentangled}
X.~Lin, J.~Cao, P.~Zhang, C.~Zhou, Z.~Li, J.~Wu, and B.~Wang, ``Disentangled deep multivariate hawkes process for learning event sequences,'' in \emph{2021 IEEE International Conference on Data Mining (ICDM)}.\hskip 1em plus 0.5em minus 0.4em\relax IEEE, 2021, pp. 360--369.

\bibitem{lin2021task}
X.~Lin, J.~Wu, C.~Zhou, S.~Pan, Y.~Cao, and B.~Wang, ``Task-adaptive neural process for user cold-start recommendation,'' in \emph{Proceedings of the Web Conference 2021}, 2021, pp. 1306--1316.

\bibitem{lin2024graph}
X.~Lin, W.~Zhang, F.~Shi, C.~Zhou, L.~Zou, X.~Zhao, D.~Yin, S.~Pan, and Y.~Cao, ``Graph neural stochastic diffusion for estimating uncertainty in node classification,'' in \emph{41st International Conference on Machine Learning (PMLR)}.\hskip 1em plus 0.5em minus 0.4em\relax MLResearchPress, 2024.

\bibitem{lin2023towards}
X.~Lin, C.~Zhou, J.~Wu, L.~Zou, S.~Pan, Y.~Cao, B.~Wang, S.~Wang, and D.~Yin, ``Towards flexible and adaptive neural process for cold-start recommendation,'' \emph{IEEE Transactions on Knowledge and Data Engineering}, vol.~36, no.~4, pp. 1815--1828, 2023.

\bibitem{10.1145/3711896.3736951}
\BIBentryALTinterwordspacing
Y.~Wu, Y.~Liu, X.~Lin, H.~Zhou, Y.~Cao, L.~Zou, Y.~Shang, and Y.~Liu, ``Faircdr: Transferring fairness and user preferences for cross-domain recommendation,'' in \emph{Proceedings of the 31st ACM SIGKDD Conference on Knowledge Discovery and Data Mining}, 2025. [Online]. Available: \url{https://doi.org/10.1145/3711896.3736951}
\BIBentrySTDinterwordspacing

\bibitem{10.1145/3711896.3737269}
\BIBentryALTinterwordspacing
S.~Zhu, M.~Li, G.~Pan, and X.~Lin, ``Ttgl: Large-scale multi-scenario universal graph learning at tiktok,'' in \emph{Proceedings of the 31st ACM SIGKDD Conference on Knowledge Discovery and Data Mining}, 2025. [Online]. Available: \url{https://doi.org/10.1145/3711896.3737269}
\BIBentrySTDinterwordspacing

\bibitem{cao2025ibpl}
Y.~Cao, F.~Shi, Q.~Yu, X.~Lin, C.~Zhou, L.~Zou, P.~Zhang, Z.~Li, and D.~Yin, ``Ibpl: information bottleneck-based prompt learning for graph out-of-distribution detection,'' \emph{Neural Networks}, vol. 188, p. 107381, 2025.

\bibitem{lin2025conformal}
X.~Lin, Y.~Cao, N.~Sun, L.~Zou, C.~Zhou, P.~Zhang, S.~Zhang, G.~Zhang, and J.~Wu, ``Conformal graph-level out-of-distribution detection with adaptive data augmentation,'' in \emph{Proceedings of the ACM on Web Conference 2025}, 2025, pp. 4755--4765.

\bibitem{yin2014temporal}
H.~Yin, B.~Cui, L.~Chen, Z.~Hu, and Z.~Huang, ``A temporal context-aware model for user behavior modeling in social media systems,'' in \emph{Proceedings of the 2014 ACM SIGMOD international conference on Management of data}, 2014, pp. 1543--1554.

\end{thebibliography}

\end{document}